\documentclass{article}
\usepackage{arxiv}
\usepackage[numbers]{natbib}
\usepackage[utf8]{inputenc} 
\usepackage[T1]{fontenc}    
\usepackage{hyperref}       
\usepackage{url}            
\usepackage{booktabs}       
\usepackage{amsfonts}       
\usepackage{nicefrac}       
\usepackage{microtype}      
\usepackage{graphicx}
\usepackage{doi}
\usepackage{amsmath}
\usepackage{amsxtra}
\usepackage{amstext}
\usepackage{amssymb}
\usepackage{mathtools}
\usepackage{amsthm}
\usepackage{acronym}
\usepackage[capitalize,noabbrev]{cleveref}

\usepackage[textsize=tiny]{todonotes}

\usepackage{tikz}
\usetikzlibrary{shapes.multipart, backgrounds,calc}

\usepackage{lipsum}
\usepackage{blindtext}
\usepackage{todonotes}
\usepackage{placeins}

\usepackage{enumitem}

\newcommand{\EE}[2][]{\mathbb{E}_{#1}\left[#2\right]} 
\newcommand{\IR}{\mathbb{R}} 

\newcommand{\ie}{\emph{i.e.}}
\newcommand{\eg}{\emph{e.g.}}

\DeclareMathOperator*{\argmax}{\mathrm{argmax}}

\newcommand{\set}[1]{\left\lbrace#1\right\rbrace}

\makeatletter
\newtheoremstyle{indented}
  {1em}
  {1em}
  {\addtolength{\@totalleftmargin}{2em}
   \addtolength{\linewidth}{-4em}
   \parshape 1 2em \linewidth \itshape}
  {}
  {\bfseries}
  {}
  {.5em}
  {}
\makeatother
\theoremstyle{indented}

\newtheorem*{question*}{Research Question}

\acrodef{HRI}{Human-Robot Interaction}
\acrodef{MDP}{Markov decision problem}
\acrodef{POMDP}{partially observable Markov decision problem}
\acrodef{PoLMDP}{policy legible Markov decision problem}
\acrodef{L-MDP}{legible Markov decision problem}
\acrodef{I-POMDP}{interactive POMDP}
\acrodef{KL}{Kullback-Leibler}
\acrodef{IRL}{inverse reinforcement learning}
\acrodef{RL}{reinforcement learning}
\acrodef{AIA}{artificial intelligent agent}
\acrodef{AI}{artificial intelligence}

\title{``Guess what I'm doing'': Extending legibility to sequential decision tasks}

\author{ \href{https://orcid.org/0000-0002-0470-0739}{\includegraphics[scale=0.06]{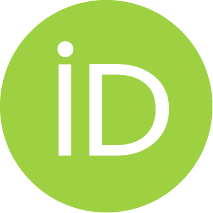}\hspace{1mm}Miguel Faria}\\
	INESC-ID \& Instituto Superior Técnico\\
	Lisbon, Portugal\\
	\texttt{miguel.faria@tecnico.ulisboa.pt}\\
	\And
	Francisco S. Melo \\
	INESC-ID \& Instituto Superior Técnico\\
	Lisbon, Portugal\\
	\texttt{fmelo@inesc-id.pt} \\
	\And
	Ana Paiva\\
	INESC-ID \& Instituto Superior Técnico\\
	Lisbon, Portugal\\
	\texttt{ana.paiva@inesc-id.pt} \\
}

\date{}

\renewcommand{\shorttitle}{``Guess what I'm doing'': Extending legibility to sequential decision tasks}

\hypersetup{
pdftitle={Acting legibly in sequential decision tasks},
pdfsubject={cs.AI, cs.Robotics},
pdfauthor={Miguel Faria, Francisco S. Melo, Ana Paiva},
pdfkeywords={planning, legibility, human-computer interaction},
}

\begin{document}
\maketitle

\begin{abstract}
	In this paper we investigate the notion of {\em legibility} in sequential decision tasks under uncertainty. Previous works that extend legibility to scenarios beyond robot motion either focus on deterministic settings or are computationally too expensive. Our proposed approach, dubbed PoL-MDP, is able to handle uncertainty while remaining computationally tractable. We establish the advantages of our approach against state-of-the-art approaches in several simulated scenarios of different complexity. We also showcase the use of our legible policies as demonstrations for an inverse reinforcement learning agent, establishing their superiority against the commonly used demonstrations based on the optimal policy. Finally, we assess the legibility of our computed policies through a user study where people are asked to infer the goal of a mobile robot following a legible policy by observing its actions.
\end{abstract}


\section{Introduction}
\label{sec:intro}

Interaction between humans and agents/robots can greatly benefit from each other's ability to reason about the others' intentions---inferring what the other is trying to do and what its objectives are. In the human-robot interaction (HRI) literature, several works have explored the communication of intentions using speech~\cite{salem15hri,paradeda19chi}, gaze~\cite{correia2017social,breazeal2005iros}, and movements~\cite{dragan2015hri, faria21roman}. In this work we address the problem of conveying intention through {\em action}, which is closely related to the aforementioned works that explore communication of intention through movement. In particular, we are interested in the notion of {\em legibility}, introduced by \citet{dragan2013hri}, that measures to what extent a user is able to infer the goal of a robot by observing a snippet of the robot's movement. 

A legible movement is characterized not by its {\em efficiency} in reaching the goal, but by its {\em distinctiveness}, i.e., by how much it is able to disambiguate the actual goal of the movement from other potential goals. In the original work of \citet{dragan2013hri}, legibility is expressed by the probability of the goal given the movement, i.e.,
\begin{equation*}
L({\rm movement})=P({\rm Goal}\mid{\rm Movement~snippet}).
\end{equation*}
Legibility has been widely explored in human-robot interaction to improve a robots' expressiveness through movement~\cite{dragan2015hri}. 

More recently, several works have extended the notion of legibility to domains other than robotic motion. The focus on improving the transparency and explainability of machine systems has been one of the main drives for the application of legibility beyond robotic motion~\cite{anjomshoae19aamas}. \citet{habibian22ral} developed a framework for legible allocation of subtasks in teams composed by robots and humans. \citet{macnally2018aamas} developed a framework for legible decision-making in deterministic scenarios. Both approaches showcase the benefits of legible behaviors when the outcome of the actions of the agent is deterministic, much as in the motion planning setting of \citet{dragan2013generating}.

\citet{miura2021roman} further extended the notion of legibility to scenarios of planning under uncertainty, introducing {\em \ac{L-MDP}}. In \acp{L-MDP}, the planning agent reasons about the observer's belief regarding about the goal of observed actions, using the multiagent framework of {\em interactive POMDP} (I-POMDPS)~\cite{gmytrasiewicz2005jair}. Unfortunately, the planning complexity of L-MDPs is similar to that of \acp{POMDP} \cite{littman98jair,madani99aaai}, making it impractical for large scale problems.

In this work we propose an alternative formulation of legibility in sequential decision making problems under uncertainty. Our framework, dubbed {\em Policy Legible MDP} (PoLMDP), avoids explicit theory of mind and, instead, defines an alternative MDP reward function that is akin to the legibility score in the original work of \citet{dragan2013generating}. Using the new reward function, our agent can compute a legible policy by solving a standard MDP, which provides a tractable alternative to L-MDPs. We show that PoLMDPs generate legible behaviors significantly faster than L-MDPs, while attaining similar levels of legibility. We also validate the legibility of the policies computed from PoLMDPs in a user study, where human users are asked to identify the goal of an agent's actions computed from a PoLMDP. Finally, we explore the impact that legible policies---computed with PoLMDP---can have when teaching other agents. Specifically, we use PoLMDP policies as demonstrations for inverse reinforcement learning agents \cite{ng00icml,ramachandran07ijcai} and show that they lead to faster learning when compared with the commonly used optimal policies.

\section{Background}
\label{sec:background}

This section introduces key concepts and notation used in the remainder of our work.

\subsection{Markov decision problems}
\label{subsec:mdps}

A {\em Markov decision problem} (MDP) is a model for sequential decision problems in stochastic environments. A MDP $M$ is defined as a tuple $\left<X, A, P, r, \gamma \right>$, with $X$ the state space; $A$ the action space; $P$ the state transition probabilities, where $P(y\mid x,a)$ indicates the probability of moving from state $x$ to state $y$ upon executing action $a$; $r:X\times A\to\IR$ is the reward function, where $r(x,a)=\EE{R_t\mid X_t=x,A_t=a}$, and $X_t$, $A_t$ and $R_t$ respectively denote the state, action and reward at time $t$; $\gamma\in[0,1)$ is a discount factor, indicating the relative importance of future rewards against present rewards.

Solving a \ac{MDP} amounts to computing an {\em optimal policy} $\pi^*$. A policy is a mapping from states to actions describing which action the agent should take in each state, and we can define the {\em value} associated with a policy as
\begin{equation*}
v^\pi(x)=\EE{\sum_{t=0}^\infty\gamma^tR_t\mid X_t=x},
\end{equation*}
where $X_t$ and $R_t$ are the state and reward at time step $t$, respectively. The optimal policy is such that $v^{\pi^*}(x)\geq v^\pi(x)$ for all $x\in X$ and all policies $\pi$. The value associated with the optimal policy is denoted $v^*$, and we define the {\em optimal $Q$-function} as
\begin{equation*}
q^*(x,a)=r(x,a)+\gamma\sum_{y\in X}P(y\mid x,a)v^*(y).
\end{equation*}
The optimal $Q$-function can be computed in polynomial time using dynamic programming \cite{papadimitriou87mor}, and the optimal policy can be computed from $q^*$ simply as $\pi^*(x)=\argmax_{a\in A}q^*(x,a)$.

\subsection{Inverse reinforcement learning}
\label{subsec:irl}

{\em Inverse reinforcement learning} is the problem of re\-co\-ve\-ring/lear\-ning a reward for an MDP given the corresponding optimal policy \cite{ng00icml}. In gradient IRL (GIRL) \cite{lopes09ecml}, the learner receives a demonstration consisting of $N$ independent state-action pairs, $D=\set{(x_n,a_n),n=1,\ldots,N}$, and computes a reward $r^*$ so as to maximize the likelihood of $D$, i.e.,
\begin{equation*}
r^*=\argmax_r\prod_{n=1}^NP(x_n,a_n\mid r),
\end{equation*}
where
\begin{equation*}
P(x_n,a_n\mid r)=\frac{1}{Z}\exp(\eta q^*_r(x_n,a_n)),
\end{equation*}
and $q^*_r$ is the optimal $Q$-function given the reward $r$, $Z$ is a normalization constant and $\eta$ is a tunable parameter. GIRL computes the reward $r^*$ using standard gradient descent.

\subsection{Legibility}
\label{subsec:legibility}

Legibility is the property that describes how readable an action or movement's objective is, and is inspired by the principle of rational action~\cite{popper1976myth}, which states that a ``{\em rational agent will act efficiently and justifiably to achieve its goals.}'' The legibility of a movement is quantified as the probability of a human assigning one specific objective, $g$, to the robot's intentions, after observing a snippet of the robot's movement, $\xi_{x_0 \to x_t}$, with $x_0$ the robot's starting pose and $x_t$ the robot's pose at time $t$. The goal inference can be defined as 
\begin{equation}
    \label{eq:legible-inference}
	\mathcal{I}_L (\xi_{x_0 \to x_t}) = \argmax_{g\in G} P(g | \xi_{x_0 \to x_t}),    
\end{equation}
where $G$ is the set of possible goals of a robot. Using Bayes' Rule
\begin{equation}
    \label{eq:legibility-target-prob}
	P(g | \xi_{x_0 \to x_t}) \propto P(\xi_{x_0 \to x_t} | g) P(g),
\end{equation}
with $P(g)$ the prior on the goals and $P(\xi_{x_0 \to x_t} | g)$ models the probability of the observed trajectory snippet being observed when the robot is moving towards goal $G$. In the original work of \citet{dragan2013hri}, the latter is expressed as a maximum entropy distribution in the form
\begin{equation*}
P(\xi_{x_0 \to x_t} | g)=\frac{\exp(-C(\xi_{x_0 \to x_t})-C(\xi_{x_t \to g}))}{\exp(-C(\xi_{x_0\to g}))},
\end{equation*}
where $C(\xi)$ is the cost associated with trajectory $\xi$.

\section{Related Work}
\label{sec:related-work}

The work we present is heavily based on the notion of legibility, as such we proceed to present the most relevant works that show the effects of applying this notion to intelligent agents. We start by exploring the applications and effects of legibility in robotics, since it was the area where the notion was first applied. Then, we move to works that have brought the use of legibility beyond robotics and to more general agent applications.

\subsection{Legibility in Robotics}
\label{subsec:legibility-robotics}
The increasingly ubiquitous presence of robots and autonomous agents in society has made paramount research on making these artificial entities clearer and transparent regarding their intentions~\citep{chakraborti2019icaps, gildert2018frontiers, alonso18frontiers}. One of the main trends of research is on improving agent transparency through implicit communication~\citep{li2017hms, huang2019ar, saunderson2019ijsr}, namely using robot motion as a means of conveying intentions~\citep{faria2016roman, huang2016hri, stulp2015iros}. Legible motions were proposed by \citet{dragan2013hri}, as a type of expressive motions, and take advantage of the Principle of Rationality~\cite{popper1976myth} combined with principles from animation, to shape robotic motion in a way that helps to disambiguate a robot's objectives. Since the introduction of legible motions, they have shown capable of improving a robot's expressiveness~\citep{dragan2015hri,faria2017iros}.

\citet{dragan2015hri} explored the impacts of legibility in one-on-one interactions between a human and a robot, when compared to motions that improve the robot's efficiency. In this work, the robot-human team had to fulfill orders in a coffee shop like scenario and the comparison between the different types of movements showed that humans found it easier and more efficient to interact with legible motions instead of other more efficient motions.

The initial works of the team of Dragan et al. have explored legibility as a property of a specific task, which can be optimized via a specific cost function. However, adapting such approach for different tasks and different classes of users is non-trivial and requires new cost functions to be designed. In an attempt to tackle this problem, \citet{busch2017ijsr} designed an approach where a robot learns to adapt its behaviour to become more legible with repeated interactions. The designed approach uses \ac{RL} to learn a behaviour model for how to interact legibly with a human. To create this model the robot needs a training phase, where it interacts with a human and learns how to perform the task and be more legible while performing it. With the developed approach, the authors showed that a robot can learn how to adapt their movements to become more legible and how to generalize this legible behaviour to different tasks without having to create a new model of interaction.

Faria et al.~\citep{faria2017iros, faria21roman} expanded to multi-party scenarios the impact of legibility in \ac{HRI}. In \citet{faria2017iros}, the authors explore the impact of applying legible motions in multi-party scenarios. The authors present a user study with a robot serving cups of water to groups of three human partners, who do not know the order through which the robot is going to serve them. The results of the study, show that using only efficient movements led to worst collaboration between the humans and the robot, than when the robot uses legible movements. When the robot focus only on using efficient movements, the humans interacting with the robot would even sometimes get confused regarding who the robot was going to serve and would get in the way of each other. Despite these initial positive results, there were some situations where the legible motions caused some confusion on the users, because the participants' perspective led them to believe that the movement was towards another person. So in a posterior work~\citep{faria21roman}, the authors used the insights from \citet{nikolaidis2016hri} to build an approach capable of generating legible motions, in multi-party scenarios, that took the different perspectives into consideration. Thus, the obtained motions improved the  legibility for all the participants, independently of their perspective over the movement. And, with a new user study conducted over M-Turk, the authors showed that this new approach was capable of generating legible motions that maintained their legibility when observed from different perspectives, leading to humans better predicting the target of the robot's motions.

Legible motions have also been used in mobile navigation. \citet{mavrogiannis18hri} have researched the use of legible motions in the planning of socially aware robotic navigation. In their work, the authors developed a framework -- \emph{Social Momentum} -- that combines an efficiency metric to drive the robot towards an intended goal with legible motions when the robot needs to avoid another agent in the same workspace. Using the Social Momentum framework, the authors showed that a robot is capable of navigating in crowded workspaces and avoid collisions. This performance was achieved by creating legible behaviours, when needing to pass by another agent, that allowed other agents to understand through which direction the robot intended to move.

\subsection{Legibility and System's Transparency}
\label{subsec:legibility-transparency}
Scepticism towards the benefits of \ac{AI} has seen a resurface in later years with prominent members of society voicing their concerns regarding the advances in the field~\cite{shermer17sa}. This growing scepticism combined with the need for \acp{AIA} to be easily understood by humans, has fuelled research in transparency of \acp{AIA}~\cite{wind2019marketing}.

Most approaches to improve \ac{AIA} transparency have focused on making more transparent technical aspects of \ac{AIA} such as the reason behind application failures~\cite{arrieta2020if, carvalho2019machine, doshi2017towards} or the inference processes of complex decision process like deep learning approaches~\cite{samek21ieee, warner2021cje, abdollahi2018transparency}. However, most of this approaches focus on the interpretability of \acp{AIA} and in making agents explainable from a user perspective, \ie~ the agent justifies its actions by presenting its reasoning after deciding, instead of making the reasoning clear while deciding. However the growing intertwining of \acp{AIA} with society, has extended the use of these agents beyond simple tools and applications we use to become peers and collaborators with whom humans interact. In this sense, \acp{AIA} need to be transparent during the entire decision process instead of just at the end of the interaction~\cite{anjomshoae19aamas}.

The ability of legible motions to shape motions to become more expressive has made this notion a good candidate to create more transparent \ac{AIA} systems~\cite{anjomshoae19aamas}. \citet{alonso18frontiers} present a mini-review of literature that gathers several works on transparency in shared autonomy workspace and explore different methods of increasing transparency. In this review, the authors conclude that legibility is one of the most common methods to increase transparency by modelling the system's behaviours.

The usefulness of legibility in making more transparent robots has led to legibility being used beyond robotic motions. \citet{habibian22ral} designed a framework to divide the subtasks of a more overarching task between humans and robots, using legible and fair allocations. The designed framework does a bilevel optimization: the first level is an optimization for the allocation of the subtasks, such that the human clearly understands what the robots' are doing and what subtasks are left for the human to do; then the second level of optimization focuses on optimizing the robots' motions to create motions that are in line with the decided allocation and thus, when observed by the human, allow him to understand what are the robots' intentions. With this work, we can observe how the notion of legibility can be applied beyond robotic motions to create allocations that are easily understood by humans.

The application of legibility to decision making has also been explored by MacNally et al. in~\citep{macnally2018aamas}. In this work, the authors formalize the decision making problem as a Goal \ac{POMDP}~\citep{kaelbling1998ai}, where the agent's goal is to choose the actions that transform an initial state belief $b_0$ into the goal belief state $b_G$. Using this formalization, the authors design a method for action selection, in deterministic scenarios, that chooses the sequence of actions that constitutes the plan that achieves $b_G$ and best disambiguates the intended goal state from other possible goal states.

The notion of legibility has also been applied to scenarios of planning under uncertainty. \citet{miura2021roman} present a formulation of legibility for \acp{MDP}, named {\em \acl{L-MDP}} (L-MDP). In \acs{L-MDP}, the agent focus on choosing, at each time step, the most optimal action that also maximizes the information transmitted to an observer about the agent's goal. To accomplish such optimal and legible behaviour, the agent reasons about the observer's belief of the agent's objective given the history of the observed agent actions. The observer's belief regarding the agent's intentions is modelled using the multiagent framework of {\em \acp{I-POMDP}}~\cite{gmytrasiewicz2005jair}. By reasoning about the observer's belief -- using this reasoning to drive the planning algorithm -- the agent can derive a {\em legible policy} that best disambiguates the agent's goal~\citep{miura2021uai}. The legible policy is obtained by iteratively updating the assumed observer's belief and with the updated belief simulate the possible actions -- using a method like UCT~\citep{kocsis2006ecml} -- to find the one that best disambiguates the agent's goal. The results of a user study, conducted by the authors, showed that the resulting legible policies are capable of better transmitting the agent's intentions than using standard optimal policies. However, the nature of \ac{L-MDP} makes its planning complexity similar to that of \ac{POMDP}, limiting its applications to small scale state spaces as the planning can become intractable in large scale state spaces.

\section{Policy Legible Markov Decision Problem}
\label{sec:polmdp}

\begin{figure}[!t]
    \centering
    \includegraphics[width=0.8\linewidth]{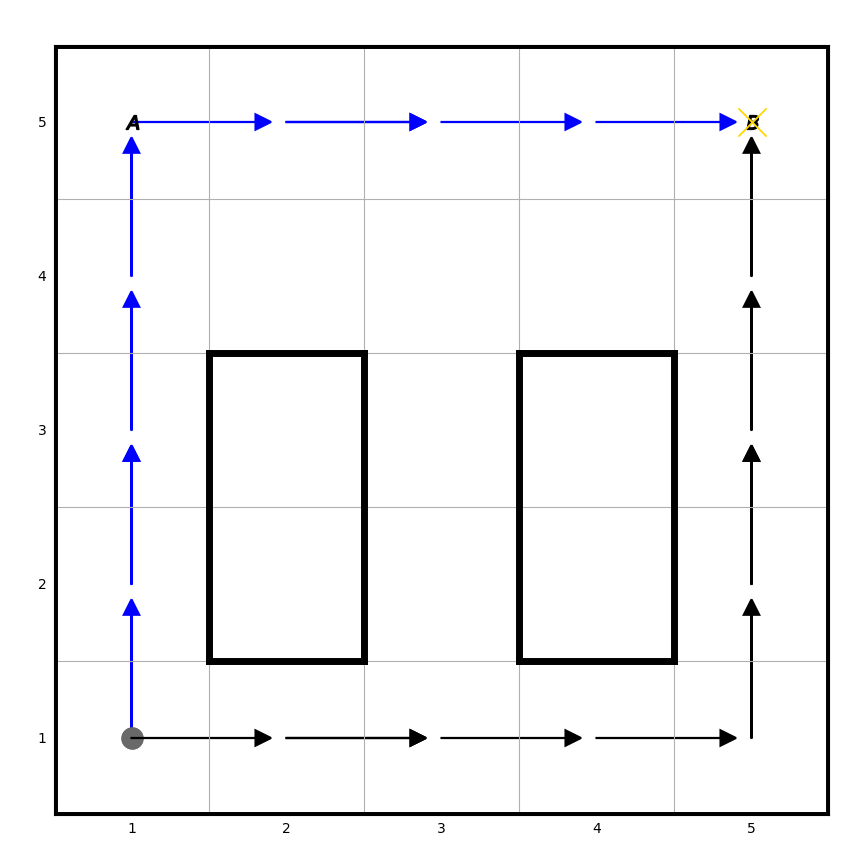}
    \caption{Example of maze-like environment with two goals, $A$ and $B$. The blue arrows indicate a possible action sequence following an optimal policy, while black arrows indicate an action sequence following a legible policy (which is also optimal).}
    \label{fig:movements-example}
\end{figure}

An optimal policy $\pi^*$ describes the most efficient way an agent can solve a \ac{MDP}. However, the optimal policy does not guarantee the decisions to be clear to an observer and may cause doubts regarding what the agent is trying to accomplish. In Fig~\ref{fig:movements-example} we can observe, in blue, a possible sequence of actions prescribed by an optimal policy, for an agent moving in a maze world scenario, towards objective $B$. We can observe that the optimal policy leads the agent to objective $B$ while going through $A$. An observer that did not know the robot's intentions could, upon observing the initial actions of the agent, confuse the robot's goal to be $A$ rather than $B$.

We propose to adapt the notion of legibility to MDPs to yield policies that offer both a solution with high expected reward and make clear the agent's current objectives. To achieve such policies we introduce \emph{\textit{\ac{PoLMDP}}}. A \ac{PoLMDP} is defined in the context of an environment with $N$ different objectives, each represented by a different reward function $r_n,n=1,\ldots,N$, and thus defining a different \ac{MDP} -- $\ac{MDP}_1, \ac{MDP}_2,\dotsc, \ac{MDP}_N$. A \ac{PoLMDP} is described as a tuple $\left< X, A, P, r_{\rm leg}, \gamma, \beta \right>$, with $X$, $A$, $P$ and $\gamma$ are as in Section~\ref{subsec:mdps}, $r_{\rm leg}$ defines the legible reward function and measures each action's legibility in each state, and $\beta$ is a non-negative constant that defines how close the legible function follows the optimal expected reward.

Following the original definition of \citet{dragan2013generating}, we define $r_{\rm reg}$ to measure the legibility of an action $a$ in state $x$ by evaluating how likely it is that $a$ is executed given that the current goal is defined by reward $r_n$, in opposition to performing the same action when trying to achieve another possible objective. In Figure~\ref{fig:movements-example} we can observe, in black, the decision sequence prescribed by a \ac{PoLMDP} policy, for an agent moving in a maze-like scenario towards objective $B$. The action sequence prescribed by \ac{PoLMDP} focus on moving the agent away from objective $A$ as soon as possible, while getting closer to $B$.

We define our legible reward function $r_{\rm leg}$ for a target reward $r_n$ as
\begin{equation}\label{eq:action-legibility}
r_{\rm leg}(x, a) = P(r_n\mid (x, a)),
\end{equation}
where
\begin{equation*}
P(r_n\mid (x, a)) \propto P((x, a)\mid r_n) P(r_n).
\end{equation*}
This is similar to Equation~\ref{eq:legibility-target-prob}. However, instead of observing a movement snippet, we observe the current state and action performed as indicators of the robot's intentions. We choose this representation because, in a \ac{MDP}, each decision step is an independent event and as such the probability of executing an action $a$ in state $x$ is not influenced by possible states and actions that preceded the current state. This way, we need only observe the current state-action pair to infer the robot's intentions. Considering a uniform distribution as the prior on the probability of observing each goal, we can simplify the previous expression as
\begin{equation}
\label{eq:legible-reward}
r_{\rm leg}(x, a) = P((x, a)\mid r_n).
\end{equation}
This probability reflects how probable is executing an action in a world state when trying to achieve one specific objective against when it tries to achieve another possible objective. We follow the same maximum-entropy principle adopted by \citet{dragan2013generating} and define $P((x, a)\mid r_n)$ as
\begin{equation}
	\label{eq:action-prob}
    P((x, a)\mid r_n) = \frac{\exp(\beta Q_n^*(x,a))}{\sum_{m=1}^N \exp(\beta Q_m^*(x,a))},
\end{equation}
with $\beta$ the parameter in the \ac{PoLMDP} description tuple and $Q_{r_n}^*$ the optimal $Q$-function for MDP$_n$. From \eqref{eq:action-prob}, an action $a$ is more legible the larger the gain of executing it in MDP$_n$ in comparison to the gain of performing $a$ for other possible goals. The policy obtained using the reward in \eqref{eq:legible-reward} promotes actions that guide the agent towards its intended goal, while increasing the agent's expressiveness.

\subsection{Relation with other approaches}

The approach taken in our \ac{PoLMDP} is closest with the L-MDP of \citet{miura2021roman}. Both approaches offer formulations of legibility for \acp{MDP}, differing on the approach taken to compute the legible policy. In the formulation of \ac{L-MDP}, the \ac{MDP} builds a reward function from a belief $b_t$ that, at each time step $t$, translates the observer's inferred goal from the actions of the agent observed up to time step $t$. The authors propose update the belief $b_t$ as follows
\begin{equation*}
	b_t \left( \theta | h_{t} \right) = P \left( \theta | s_{t-1}, a_{t-1}, s_t, b_t \right)
\end{equation*}
\begin{equation*}
	 = \frac{\hat{T} \left( s_{t-1}, a_{t-1}, s_t | \theta \right) \hat{\pi} \left( s_{t-1}, a_{t-1}| \theta \right) b_{t - 1} \left( \theta \right)}{\sum_{\theta^{\prime}} \hat{T} \left( s_{t-1}, a_{t-1}, s_t | \theta^{\prime}\right) \hat{\pi} \left( s_{t-1}, a_{t-1}| \theta^{\prime} \right) b_{t - 1} \left( \theta^{\prime} \right)}, 
\end{equation*}
with $\theta$ the agent's goal, $\hat{T}$ is the assumed transition for a given $\theta$ and $\hat{\pi}$ the agent's policy for a given $\theta$. The reward is computed from either the Euclidean distance or KL-divergence between the observer's (estimated) belief, $b_t$, and the belief $b^*$ translating the correct goal
\begin{equation*}
	dist(b_t, b^*) = D_{KL} (b_t \,||\, b^*) = \sum_{\theta \in \Theta} b_t(\theta) log \frac{b_t(\theta)}{b^*(\theta)}.
\end{equation*}
The dependence of the reward on $b_t$ introduces a dependence on the history of the process, rendering L-MDPs not amenable to the use of standard MDP (or POMDP) solution techniques. 

Our approach circumvents the dependence on the history by pre-computing $b ( \theta )$ using \eqref{eq:action-prob} and can be obtained from that of \citet{miura2021roman} by instead considering the distance between beliefs given by the \emph{Total Variation} (TV) distance
\begin{equation*}
    dist(b_t,b^*) = \frac{1}{2} D_{TV}(b_t \,||\, b^*) = \frac{1}{2} \sum_{\theta \in \Theta} |b_t(\theta) - b^*(\theta) -1 |,
\end{equation*}
where $b^*$ is an indicator function for the agent's goal. The KL-divergence, used by Miura et al., is a special case of $f$-divergence with $f(x)=x \log x$, while the TV-distance we use in \ac{PoLMDP} is a special case of  $f$-divergence with $f(x) = |x - 1|$.

\section{Experimental Evaluation}
\label{sec:eval}

In this section we present the evaluation of our framework. Our evaluation aimed at answering three questions:
\begin{enumerate}
	\item how our framework performs compared to the framework of \ac{L-MDP} proposed by Miura et al.~\cite{miura2021roman}, thus understanding how \ac{PoLMDP} stands between other types of legible sequential decision making frameworks;
	\item whether the use of \ac{PoLMDP} generated samples teach an \ac{IRL} agent better, than using samples generated using a standard optimal policy;
	\item in an interaction with human users, if a robot using a legible policy generated by \ac{PoLMDP} convey its intentions faster than using a standard optimal policy.
\end{enumerate}

To answer these three questions we divided the evaluation in three steps.

We start by explaining the scenario used in the three evaluation steps, detailing the motivation for using the chosen scenario and the main decisions behind the adaptations in the scenario. After explaining the scenario, we present the three evaluations performed.

\subsection{Evaluation Scenario}
\label{subsec:scenario}

When deciding on a scenario to use in our evaluations, we had two main concerns in mind. The first concern was that the scenario used could be reused in the three evaluations, specially knowing that we had both comparisons between legible frameworks in simulation scenarios and comparisons between agent decision approaches in user studies. Thus, the scenario had to fully allow comparison between the performance of different frameworks for legible acting but also be a scenario that would occur in a natural interaction between humans and an agent system. The second concern was that the environment's dynamics would not cause an impact in the legibility of actions, this because during an interaction between a human and an agent, the human only observes the outcome of the actions and not the action itself. So, in the case of a failed action, the failure must not influence the legibility of the robot, otherwise we add artifacts and errors to the legibility that cannot be measured.

Having these two concerns in mind, we decided to use a Maze World scenario, where a robot had to navigate and reach one of various coloured areas scattered in the maze. These areas would serve as the different goals the robot could be aiming to accomplish. The robot had available five different actions: moving up, moving down, moving left, moving right and no operation. When moving in the maze, the robot had a 15\% chance of failing the action and staying in the same location.

The Maze World scenario chosen is a scenario widely used to compare agents and robotics frameworks and is interesting because it serves as the base to model different real world scenarios, \eg \, the plan of a building during a search and rescue mission or the floor plan of a warehouse with robots retrieving different items to deliver. So, this is a scenario that can be reused in the three planned evaluations respecting our first concern. Secondly, the 15\% chance of failing and staying in the same place may seem to simplify the scenario, but it introduces stochasticity and does not impact the legibility of the robot in the case of a failure. Other options would be for the robot to veer off course in a failure, but by veering off course we would be introducing artifacts that could impact the legibility in unforeseen ways and thus impact the measured levels of legibility.

\subsection{Comparison with similar frameworks}
\label{subsec:framework-comparison}

The comparison of our framework with other similar frameworks is an important aspect so we can understand what are the advantages our framework presents. Thus, we compared our framework with Miura's \ac{L-MDP}~\cite{miura2021roman}. We only compared with this framework because it is the only framework that proposes to solve the problem of legible decision making using stochastic environments.

\subsubsection{Setup}
\label{subsubsec:framework-comparison-setup}

Our comparison used a maze world formulation and explored how each framework performed when we scaled:
\begin{itemize}
    \item the number of possible goals, keeping constant the number of states;
    \item the number of states in the world, keeping the number of goals constant.
\end{itemize}

The measures we used to compare the frameworks were: the average time taken, by each framework, to find a sequence of decisions from the initial state to the given goal; the average legibility value for each sequence using the legible function of \ac{PoLMDP} and using the legible function of Miura's \ac{L-MDP}.

To compare the performance of the two frameworks with varying number of goals, we tested both frameworks on a 25x25 maze world with the number of possible goals varying between 3 and 10 possible goals, making up 7 different world configurations. 


Regarding the states scalability test, we designed multiple maze world configurations where we varied the number of states. All mazes had 6 possible goals, except for the smallest maze which size could only accommodate 3 goals. The smallest maze was taken from the paper where Miura presents the \ac{L-MDP} framework and has 40 states, a 5 rows by 8 columns maze. The rest of the mazes had the following dimensions: 100 states (10x10), 625 states (25x25), 1600 states (40x40), 2500 states (50x50), 3600 states (60x60) and 5625 (75x75). Again totalling 7 different world configurations. 

For each test we sampled 250 pairs of initial state and goal for each world configuration, with the only requirement being that the initial state was different than the goal state.

With these tests we aim at comparing the performance of the two frameworks under different scalability conditions, namely the impact on the average time taken to obtain a solution and the average quality of the solutions.

After sampling the 250 pairs for each test and each world configuration, we ran those pairs through each framework, limiting their execution time to a maximum of 2 hours for each test pair. If in those 2 hours, a framework could not give a solution, we would mark that test pair as a failure.

\subsubsection{Results}
\label{subsubsec:framework-comparison-results}

\begin{figure}[t]
	\centering
	\includegraphics[width=0.8\linewidth]{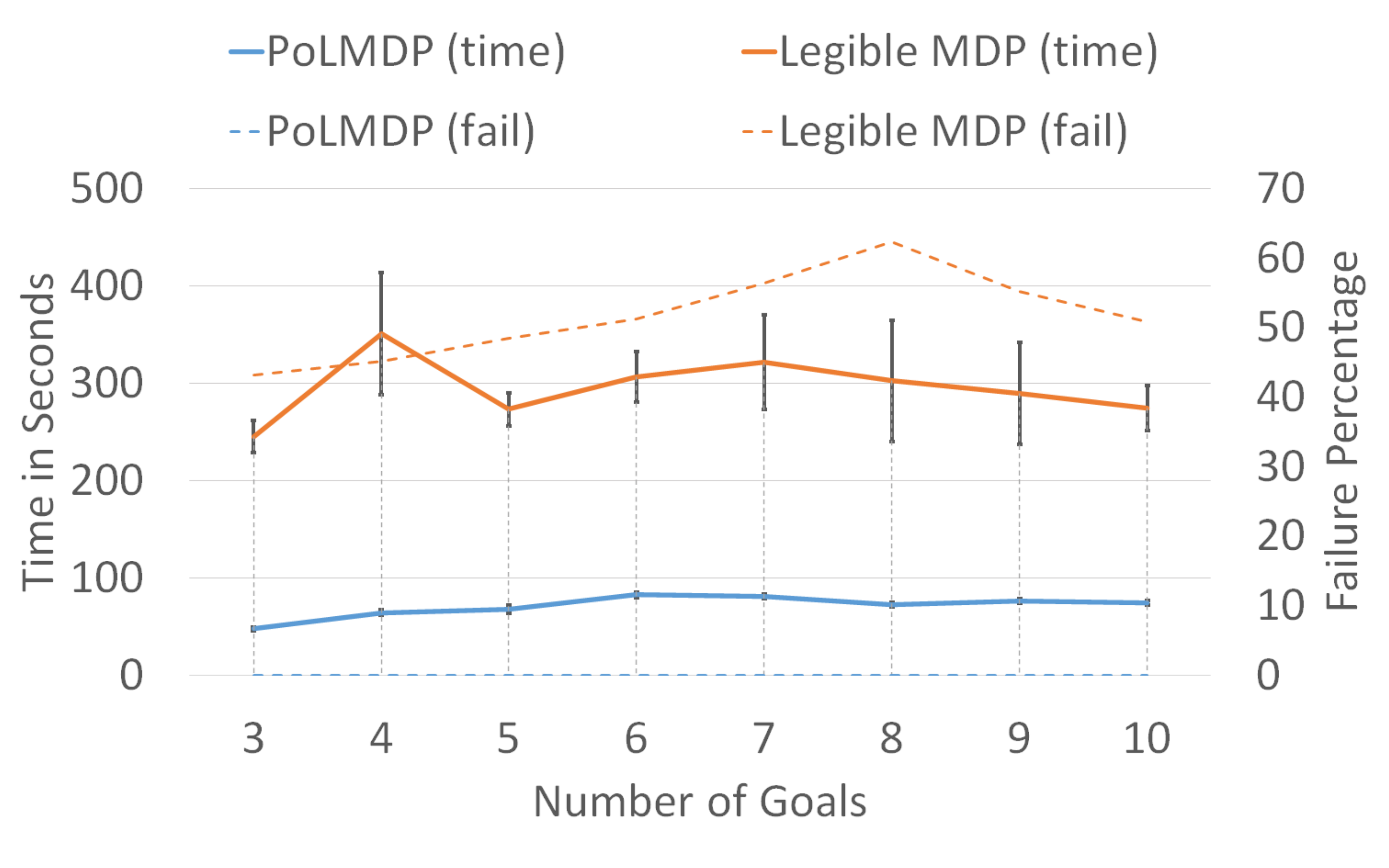}
	\caption{Results for the time performance comparison between the \ac{PoLMDP} framework against Miura's Legible \ac{MDP}, when we vary the number of possible goals in a mazeworld like scenario. In continuous lines we show the average times, and, in dashed lines, the percentage of failed tests.}
	\label{fig:goals-eval-time}
\end{figure}

\begin{figure}[t]
	\centering
	\includegraphics[width=0.8\linewidth]{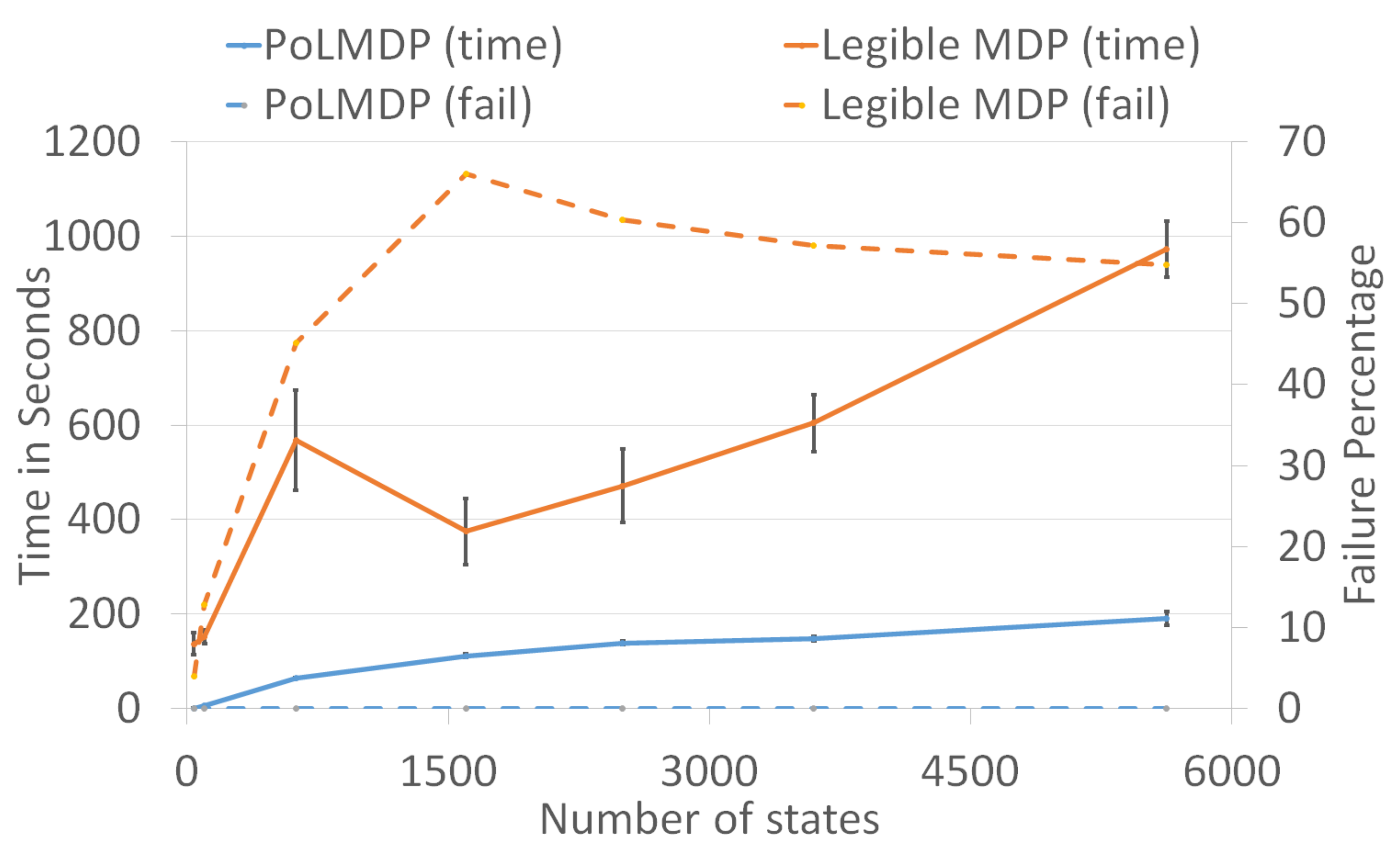}
	\caption{Results for the time performance comparison between the \ac{PoLMDP} framework against Miura's Legible \ac{MDP}, when we vary the number states in the mazeworld scenario. In continuous lines we show the average times, and, in dashed lines, the percentage of failed tests.}
	\label{fig:states-eval-time}
\end{figure}

With all the testing pairs ran, we had to do a pre-processing of the results to guarantee a balance dataset of the results. This step was needed because of the disparity of failed tests between the two frameworks, as shown in dotted lines in Figures~\ref{fig:goals-eval-time} and~\ref{fig:states-eval-time}, which caused to have an unbalanced number of test results for our \ac{PoLMDP} and for Miura's \ac{L-MDP}. Thus, to balance the datasets, we decided to use only 100 samples for each testing condition, which was the highest common number of samples between all testing conditions, \ie we used 100 samples of each of the 7 world configurations for each of the testing conditions.

Given that we used a time constraint to determine test failure, we used the same time criteria to balance the dataset. Thus, for each world configuration in both the state scaling and goal scaling conditions, we ordered the results by time the time taken to obtain a solution and, for each world configuration, we removed the result entries that presented higher execution time. This way we reduced the amount of bias introduced by using the same criteria as before.

After balancing the results for each configuration, we extracted 100 test results for each configuration. For the configurations that had more than 100 samples, we randomly selected 100 samples for each configuration, considering a sample the results corresponding to same pair of initial state and goal for both frameworks. For the configurations with only 100 samples, we did no random sampling and used the available 100 samples.


For the goals scalability test, the percentage of failures is shown in Figure~\ref{fig:goals-eval-time} in dashed lines, while Figure~\ref{fig:states-eval-time}, in dashed lines, shows the percentages of failures for the states scalability test.

\begin{figure}[t]
	\centering
	\includegraphics[width=0.8\linewidth]{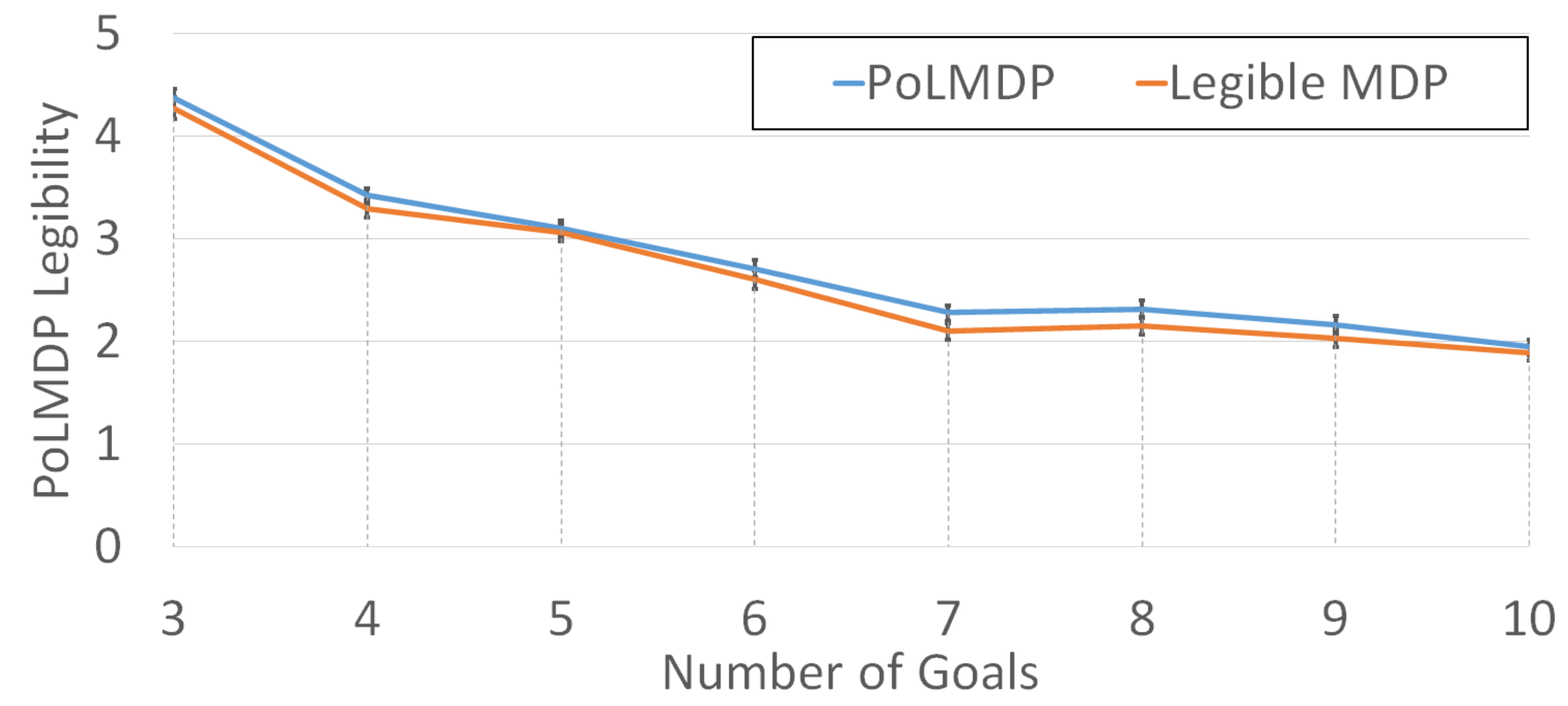}
	\caption{Results for the \ac{PoLMDP} legibility metric performance comparison between the \ac{PoLMDP} framework against Miura's Legible \ac{MDP}, when we vary the number of possible goals in a mazeworld like scenario.}
	\label{fig:goals-eval-policy}
\end{figure}
\begin{figure}[t]
	\centering
	\includegraphics[width=0.8\linewidth]{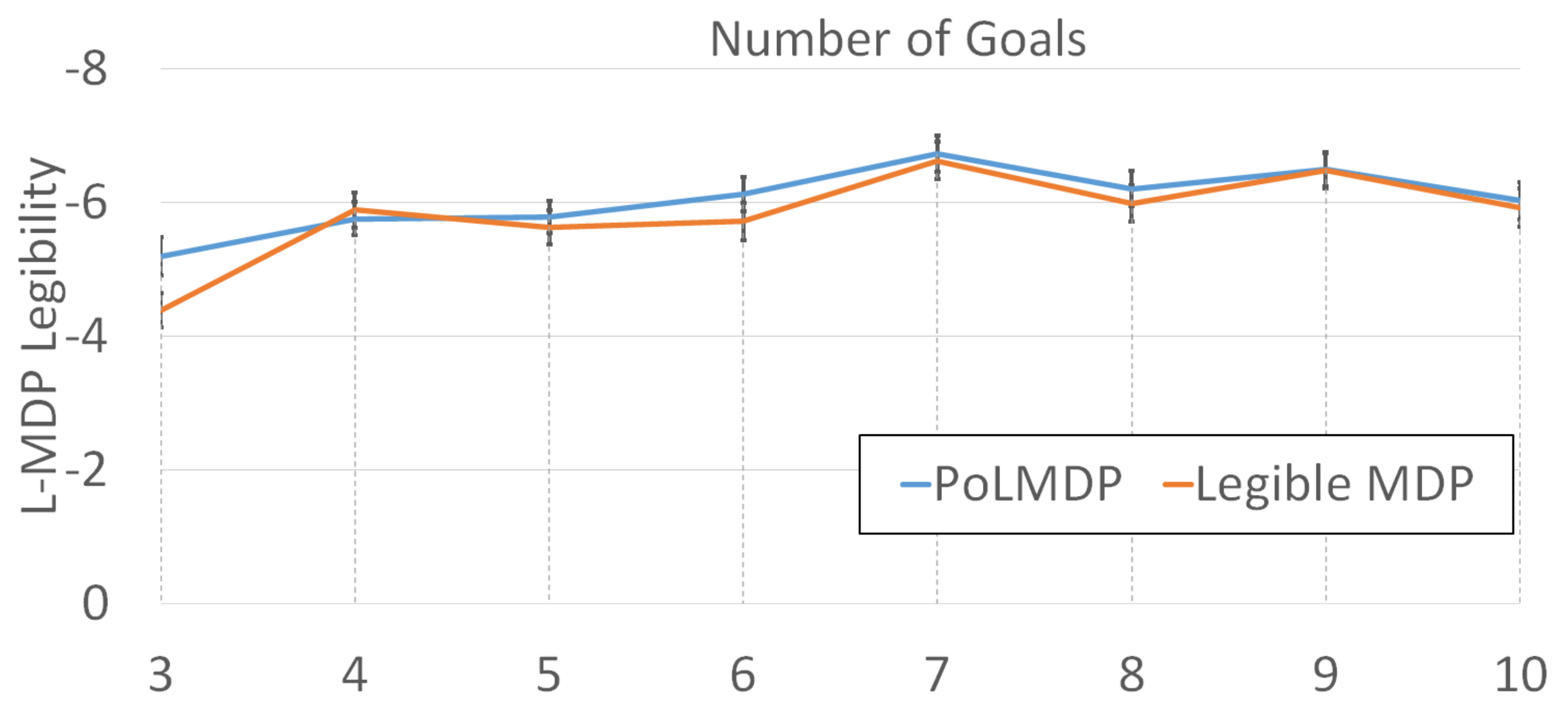}
	\caption{Results for the Miura's Legible \ac{MDP} legibility metric performance comparison between the \ac{PoLMDP} framework against Miura's Legible \ac{MDP}, when we vary the number of possible goals in a mazeworld like scenario.}
	\label{fig:goals-eval-miura}
\end{figure}
\begin{figure}[t]
	\centering
	\includegraphics[width=0.8\linewidth]{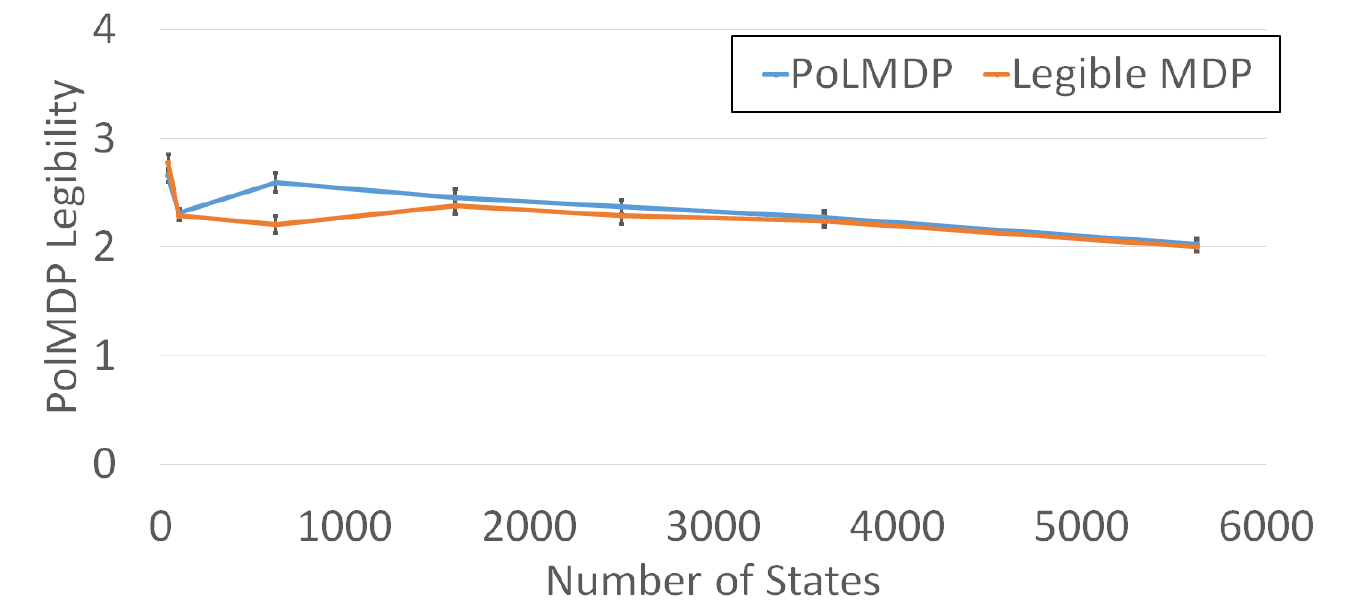}
	\caption{Results for the \ac{PoLMDP} legibility metric performance comparison between the \ac{PoLMDP} framework against Miura's Legible \ac{MDP}, when we vary the number states in the mazeworld scenario.}
	\label{fig:states-eval-policy}
\end{figure}
\begin{figure}[t]
	\centering
	\includegraphics[width=0.8\linewidth]{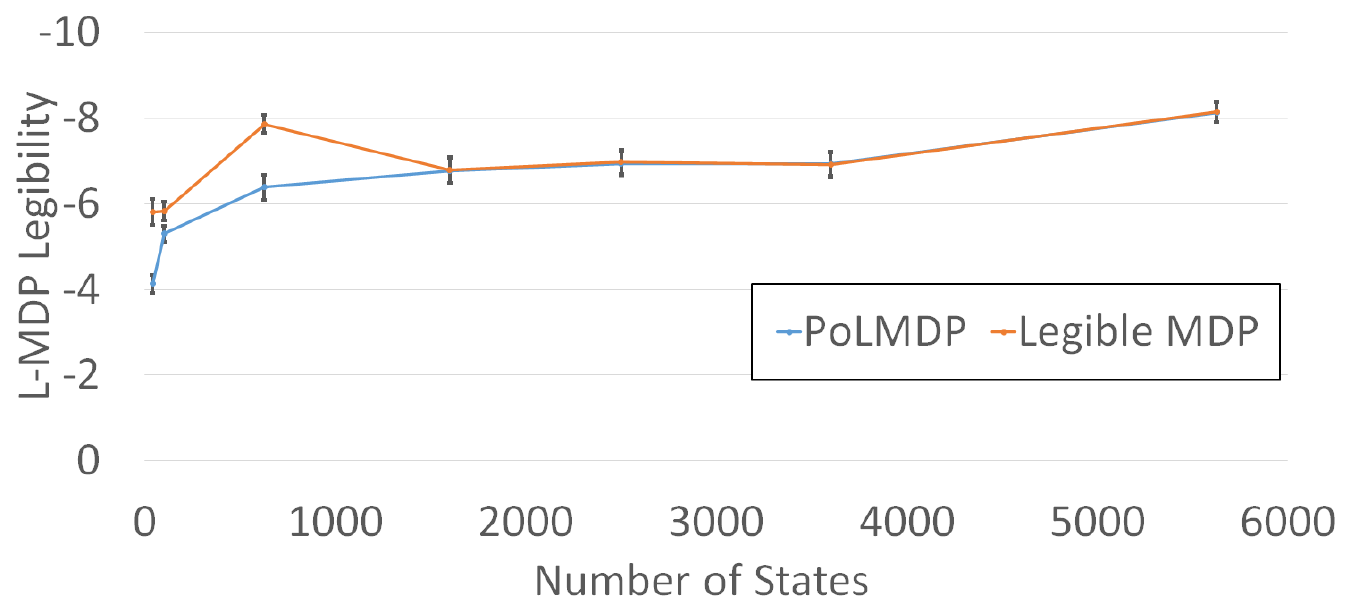}
	\caption{Results for the Miura's Legible \ac{MDP} legibility metric performance comparison between the \ac{PoLMDP} framework against Miura's Legible \ac{MDP}, when we vary the number states in the mazeworld  scenario.}
	\label{fig:states-eval-miura}
\end{figure}

Regarding the evaluation for the average time taken to give a solution and the legibility performance according to each of the two legibility metrics used, Figures~\ref{fig:goals-eval-policy},~\ref{fig:goals-eval-miura} and~\ref{fig:goals-eval-time} show the results for the goals scalability and Figures~\ref{fig:states-eval-policy},~\ref{fig:states-eval-miura} and~\ref{fig:states-eval-time} show the evaluation results for the state scalability test. Figures~\ref{fig:goals-eval-policy} and Figure~\ref{fig:states-eval-policy} show the average legibility, according to the \ac{PoLMDP}'s legibility function, of the obtained solutions for each framework; Figures~\ref{fig:goals-eval-miura} and Figure~\ref{fig:states-eval-miura} show the average legibility, according to the \ac{L-MDP}'s legibility function, of the obtained solutions for each framework; and Figures~\ref{fig:goals-eval-time} and Figure~\ref{fig:states-eval-time} show the average time each framework needed to obtain a possible reward.




\subsubsection{Discussion}
\label{subsubsec:framework-comparison-discussion}

The results of the first evaluation show interesting results. The first result that deserves analysis is the percentage of tests that failed. The \ac{L-MDP}s had an average of 40-60\% fail rate in most of the tests in both scalability tests. The only cases where such did not occur where on the tests with world configurations with small state spaces ($\leq100$ states), where the fail rate was between 4\% and 13\%. On the other hand, looking to the results of \ac{PoLMDP}, we can observe that the fail rate was always at 0\%, that means \ac{PoLMDP} was capable of always finding a solution in under 2 hours, no matter how big the number of states or the number of goals presented. This is an extremely interesting fact, because when we use sequential decision systems in robots, we need them to give a solution quickly and the \ac{PoLMDP} offers that capacity. These results are closely related with the results obtained for the average time to find a solution, where again \ac{PoLMDP} outperformed \ac{L-MDP}; the analysis of the average time also yields another interesting fact, with an increasing number of states the \ac{PoLMDP}'s average time to find a solution increases at much lower rate than that of \ac{L-MDP}, hinting at the \ac{PoLMDP} being more resistant to changes in the number of states.

Regarding the results of the legibility metrics, the two frameworks did not show a significant difference on the results obtained for the scalability of goals. Both frameworks performed better according to their respective metrics, but without significant differences.

\subsection{Evaluation with an IRL agent}
\label{subsec:irl-evaluation}

\ac{IRL} agents use examples given by an expert to try and learn the expert's underlying reward function to solve a given sequential decision problem. To that extent, the example samples given to an \ac{IRL} agent must clearly show the expert's preferences and best practices to solve the problem. In a single objective scenario, simply presenting examples of the best practices allows an \ac{IRL} agent to learn the reward function being used; however, in a scenario with multiple possible objectives that may not be the case, because the best practices to achieve one of the goals may be the same to achieve a set of other goals. With that in mind, we decided to evaluate if using legible examples given by our \ac{PoLMDP} approach allow an \ac{IRL} agent to learn the underlying reward function better, than using simply examples of the best actions to take, in a multiple objectives scenario.

\subsubsection{Setup}
\label{subsubsec:irl-setup}

To understand the best approach to teach an \ac{IRL} agent, we did two different tests. The first test aimed at understanding how the two approaches fared when teaching an \ac{IRL} agent a complete sequence of decisions from beginning to end, resembling how a human expert teaches the complete process of solving a problem. This test explores how the two approaches teach an \ac{IRL} agent with examples that are logically expected to follow previous ones. The second test aimed at exploring how well the two frameworks teach an \ac{IRL} agent the best actions in situations without a necessary connection between them. This test resembles how an expert would teach fringe situations in solving a problem where, instead of showing the entire process, the expert would focus on showing how to act in specific situations. With this test, we explore how well the two approaches teach an \ac{IRL} agent using examples that do not necessarily follow previous ones.

In this evaluation we used 4 different configurations of 10x10 mazes. The 4 configurations varied both on the position of the possible goal locations and the configuration of the walls in the mazes, however in all the configurations there were 6 possible goal locations the robot could reach. 
%

For the comparison of sequential decisions, we sampled 250 pairs of initial positions in the mazes, for each of the 4 world configurations. Then, using either a policy generated by \ac{PoLMDP} or the optimal policy, we obtained, for each initial position, ten trajectories of 20 steps each between that initial position and each of the possible goals. With all the trajectories generated, we then took each trajectory and sequentially gave more examples of the same trajectory to an \ac{IRL} agent. Each time we supplied a new example, we asked the \ac{IRL} agent to predict what was the robot's goal, using the samples of the trajectory seen so far, registering if the prediction was correct or incorrect. We used a different instantiation of an \ac{IRL} agent for trajectories generated with a \ac{PoLMDP} policy and trajectories generated with an optimal policy.

For the comparison of non-logically connected examples, we sampled 250 sets with 20 random states each, for each of the 4 world configurations. Then we obtained the action prescribed, using either the \ac{PoLMDP} policy or the optimal policy, for each random state to move towards each one of the possible goals. After finishing this process, we took each set of 20 state-action pairs and sequentially gave each pair to an \ac{IRL} agent, asking the agent to predict, with each new pair, the goal most probable for the robot. Each time we got a prediction, we compared it to the expected prediction and marked either as correct or incorrect. We used a different instantiation of an \ac{IRL} agent for trajectories generated with a \ac{PoLMDP} policy and trajectories generated with an optimal policy.

\subsubsection{Results}
\label{subsubsec:irl-results}

After running the evaluation on both testing conditions, we aggregated the results and obtained the average ratio of correct predictions, depending on the approach used and on the number of samples given to the \ac{IRL} agent.

\begin{figure}[t]
	\centering
	\includegraphics[width=0.9\linewidth]{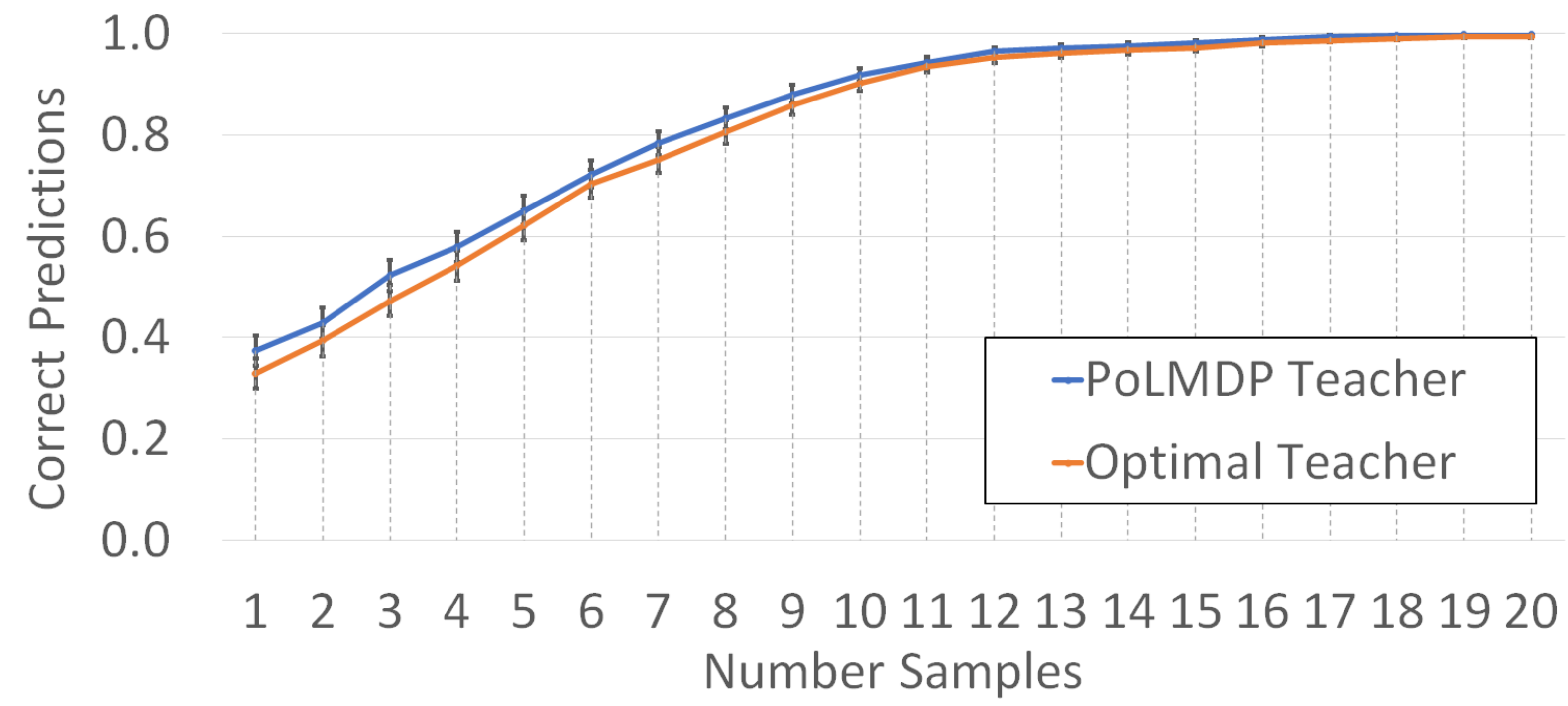}
	\caption{Ratio of correct predictions depending on the number of examples shown to an \ac{IRL} agent, with error bars. These results pertain to the condition where the samples formed a complete sequence of decisions.}
	\label{fig:irl-trajectory}
\end{figure}

The results for the test with complete decision sequences can be observed in Figure~\ref{fig:irl-trajectory}. The results show that both approaches allowed the agent to learn at a similar rate, although the samples from a \ac{PoLMDP} policy appear to allow the \ac{IRL} agent to correctly predict the teacher's goal more frequently when less samples were shown. However these results do not show a significant difference between the two approaches when the examples used are related to each other.

\begin{figure}[t]
	\centering
	\includegraphics[width=0.9\linewidth]{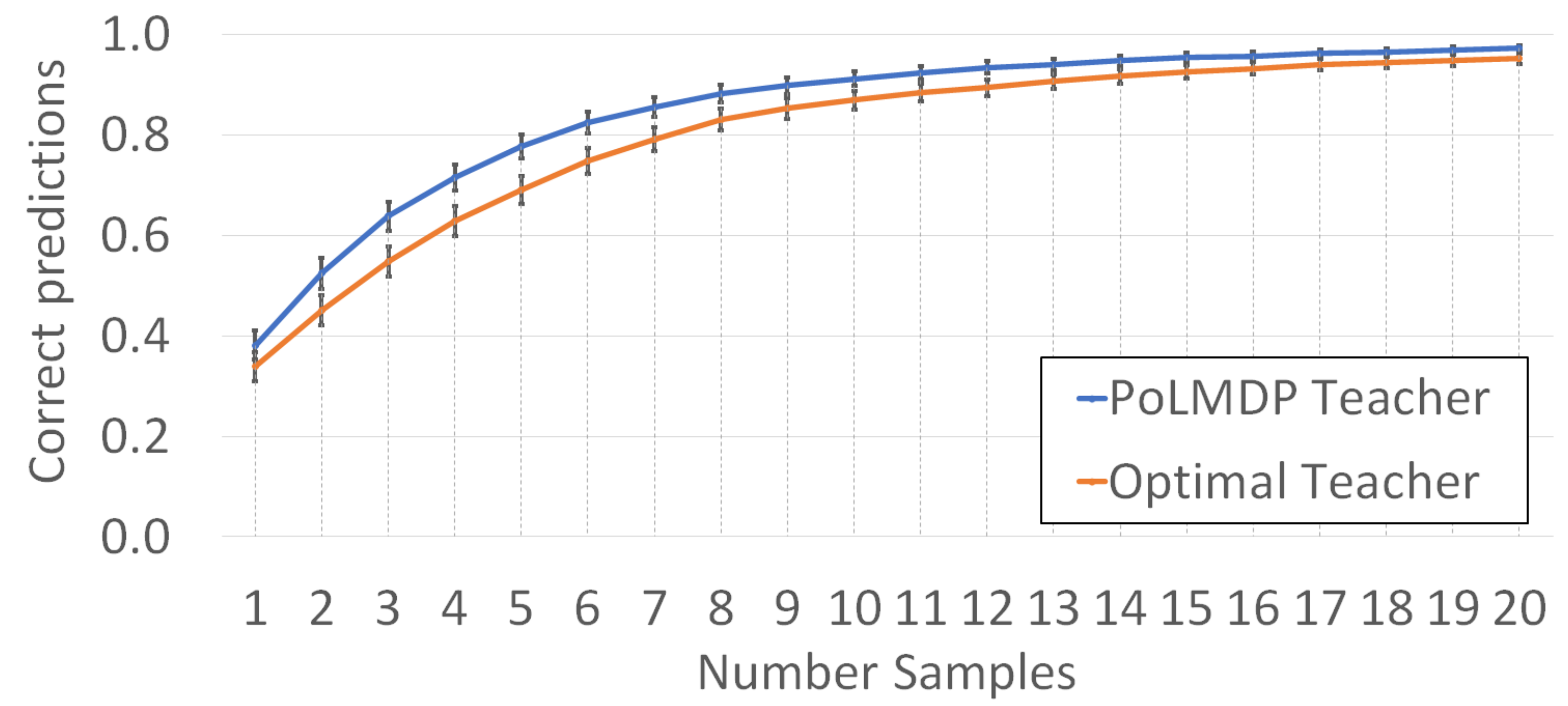}
	\caption{Ratio of correct predictions depending on the number of examples shown to an \ac{IRL} agent, with error bars. These results pertain to the condition where the samples had no specific correlation between each other.}
	\label{fig:irl-samples}
\end{figure}

The results for the test with non-logically connected examples can be observed in Figure~\ref{fig:irl-samples}. The results show that both \ac{IRL} agents learned the teacher's intentions after 20 examples, but the two approaches have shown differences in performance. When using examples from a \ac{PoLMDP} policy, the \ac{IRL} agent learned the teacher's intentions at a significantly faster pace than when the teacher used examples from an optimal policy. When analysing the results, we observe that, when using a \ac{PoLMDP} policy, with only 5 examples the \ac{IRL} agent correctly predicted the learner's intentions in ~80\% of the trials. The same performance was only achieved by an \ac{IRL} agent using an optimal policy after observing 7 examples. Only after 12 samples the performance of an \ac{IRL} agent becomes similar using either approaches, with an agent using a \ac{PoLMDP} policy showing better learning before then.

\subsubsection{Discussion}
\label{subsubsec:irl-discussion}

The results of both tests show that the use of \ac{PoLMDP} policies allows an \ac{IRL} agent to learn the underlying reward function and the teacher's intentions faster than using a standard optimal policy. This difference is more distinct in the test with examples without a necessary relation between them. In this test we observed that the \ac{IRL} learner paired with the \ac{PoLMDP} teacher continually had a higher correct prediction ratio than the learner paired with the optimal teacher.

The results for the test with a complete sequence of decisions do not show a significant difference between both approaches. This was not surprising, because in this test both approaches had to give a sequence of decisions logically connected with each other and thus the examples sequentially drove the learner towards the correct goal. However, the results show a slight trend for the performance of a learner paired with a \ac{PoLMDP} teacher to be better, which is in line with the principles of legibility that aim at making actions easier to read and understand the intentions behind them.

Overall, the results of this evaluation point to the \ac{PoLMDP} framework allow for an \ac{IRL} agent to learn the teacher's underlying reward function faster.

\subsection{User Study}
\label{subsec:user-study}

The user study is an important aspect of our evaluation because this framework is meant to improve the interaction between robots or other autonomous agents and humans. Thus, to correctly infer the impact \ac{PoLMDP} has on possible human users, we need to evaluate if a robot using a \ac{PoLMDP} policy is better at conveying intentions than a robot using a policy that maximizes the underlying reward function.

\subsubsection{Setup}
\label{subsubsec:study-setup}

We conducted a user study on the Prolific\footnote{https://prolific.co/} platform, an online platform to conduct academic research and data collection. Our study aimed at comparing the performance of our \ac{PoLMDP} framework in conveying a robot's internal goals and explored the question:
\begin{center}
    \emph{``Does the use o \ac{PoLMDP} generated policies lead a more informative robot decision making?''}
\end{center}

To support our exploration of the problem and answering the research question, we postulate the following working hypotheses:
\begin{enumerate}[label={\bf H\arabic*}]
    \item \emph{Participants will understand better the robot's intentions, when paired with a \ac{PoLMDP} policy than when paired with an optimal policy}.
    \item \emph{Participants will understand quicker and more confidently the robot's intentions, when paired with a \ac{PoLMDP} policy than when paired with an optimal policy}.
\end{enumerate}

\begin{figure}[t]
	\centering
	\includegraphics[width=0.7\linewidth]{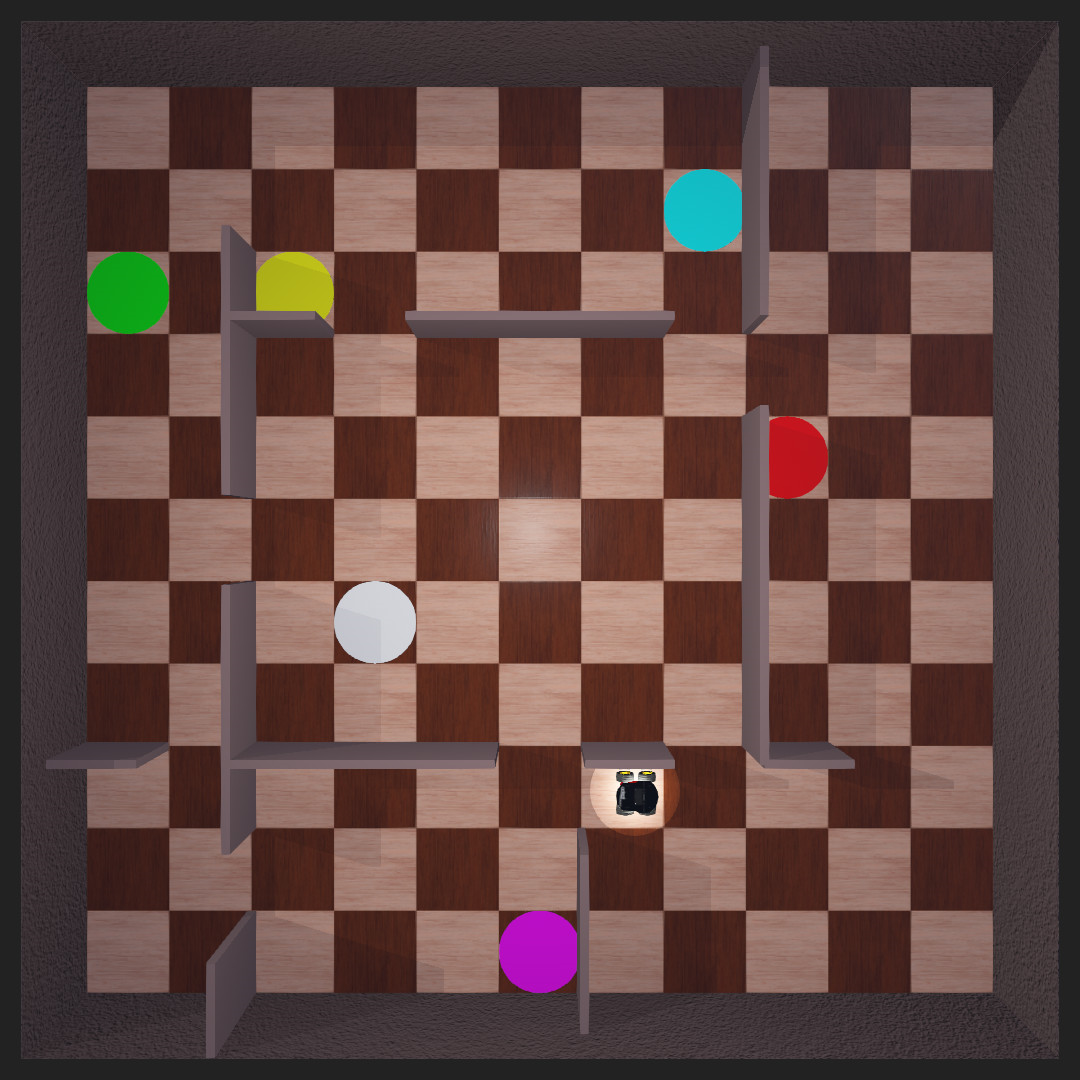}
	\caption{Still of a possible video the participants would watch. Each of the coloured circles is a possible location the robot could moving towards and the participants had to predict the correct one as soon as possible.}
	\label{fig:study-example}
\end{figure}

Thus we designed an online study, disguised as a guessing game, where the participants had to correctly predict where the robot was moving towards. Our study followed a between-subjects design, with each group being a different type o policy -- \ac{PoLMDP} policy or optimal policy. Each participant would observe 10 small videos of a robot moving, in a maze world scenario, to one of 6 differently coloured areas. Figure~\ref{fig:study-example} shows a still from one of the possible videos the participants would watch. The participants had to correctly predict which of the coloured areas was the robot's objective and as fast as possible. For each video, the participants had a play and stop button to control how much of the video to watch and when they felt they knew the objective they would stop the video, choose what coloured circle was the robot's objective and rate how confident they were in the prediction.

At the end of the study, the participants were presented with a score on how well they performed depending on how fast they correctly predicted the robot's objectives.

Before beginning the study, the participants had to read a consent form and choose if they consented to the study. After that they answered some demographic questions about their age, gender, nationality, degree of education, occupation and familiarity with robots.

\subsubsection{Results}
\label{subsubsec:study-results}

We recruited 150 participants through the Prolific platform, with ~66\% from Europe, ~19\% from Africa, ~11\% from Northern America and Mexico and the remaining ~3\% from South America and the Middle East. The participants average age was between 18 and 29 years old, but their ages varied from 18 to 70 years old. Their gender distribution was 59\% female, ~38\% male and less than 3\% identified as non-binary. Finally, ~88\% reported to have had little or no interaction with robots in their lives, ~9\% reported to interact occasionally with robots and less than 3\% interact frequently with robots. 

Besides the demographic data, for each participant we measured the average time taken to predict the robot's objective and the percentage of correct predictions. We also measured the self-disclosed rating of confidence in the predictions. Since each participant answered to 10 different videos and the videos were randomly sampled from a pool of 37 videos for each condition, we measured the participants' answers for each video presented.

To measure the average time taken to predict the robot's objective, we recorded the timestamp on the video when the participant stopped the video before making a prediction and the full length of the video. Then, if a participant correctly predicted the robot's objective, the participant's time would be the recorded timestamp, if the prediction was wrong the prediction time would be the full length of the video. Regarding the rating in confidence, each participant was asked to rate their confidence in a 7 point Likert scale after issuing a prediction in each video.

Before we started the analysis of the results, we conducted a normality test on the three measures used. This test reported that the answers obtained did not follow a normal distribution, so all analyses use non-parametric tests or tests that do not assume normality of answers.

\begin{figure}[t]
	\centering
	\includegraphics[width=0.7\linewidth]{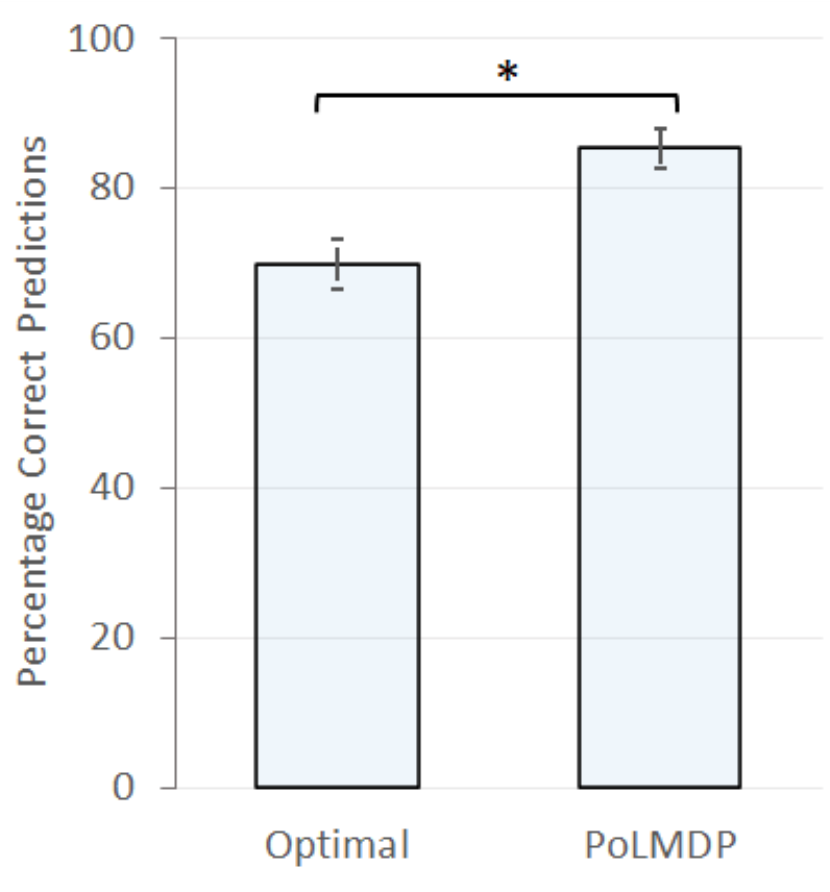}
	\caption{Percentage, with 95\% confidence interval error bars, of participants that correctly predicted the robot's objectives, according to the type of policy used. (*$p < 0.05$)}
	\label{fig:study-correct}
\end{figure}

The analysis of the number of correct predictions yielded that participants paired with \ac{PoLMDP} correctly predict the robot's objective in ~85\% of the predictions, while those paired with the optimal condition correctly predicted the objective in ~70\% of the predictions. We conducted a Pearson's Chi-Square test, $\chi^2 (1) = 48.864, p < 0.001$, showed that the difference in percentages was significant, supporting our \textbf{H1} hypothesis. Figure~\ref{fig:study-correct} shows the results for the percentage of correct answers according to the type of policy used.

\begin{figure}[t]
	\centering
	\includegraphics[width=0.7\linewidth]{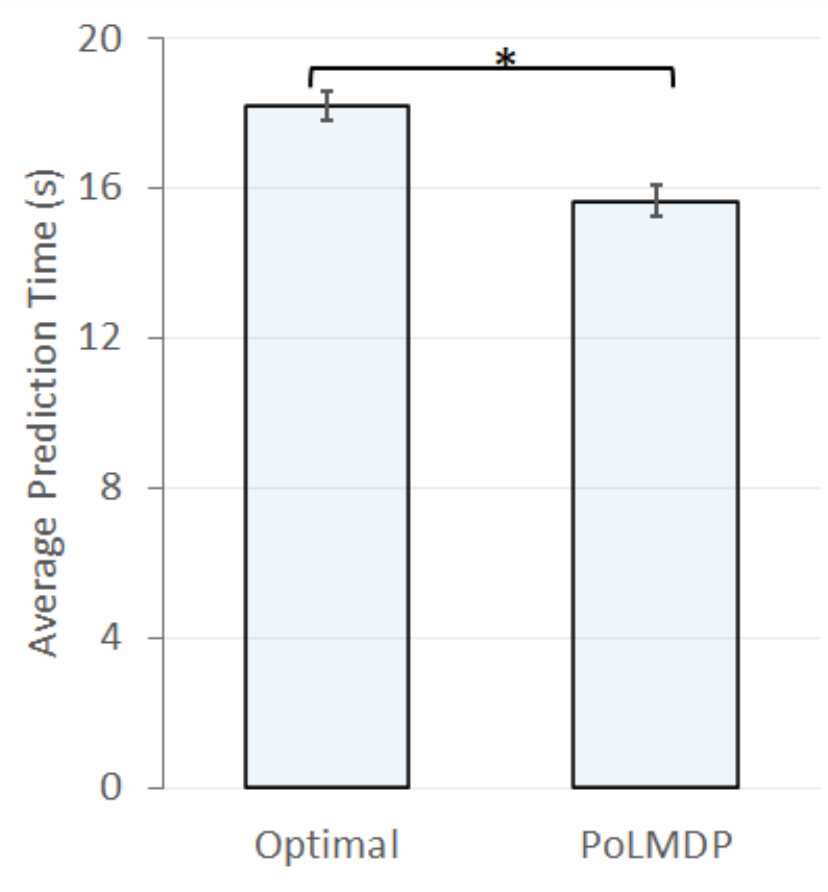}
	\caption{Average time, with 95\% confidence interval error bars, to correctly predict the robot's objective, according to the type of policy used. (*$p < 0.05$)}
	\label{fig:study-time}
\end{figure}

The analysis of the average time to correctly predict the objective, showed that on average participants paired with the \ac{PoLMDP} condition took 15.67 seconds to correctly predict the robot's objective, while participants paired with the optimal condition took 18.22 seconds. We conducted a Mann-Whitney test that showed the difference was significant, $U = 177976, p < 0.001$, thus supporting our hypothesis \textbf{H2}. Figure~\ref{fig:study-time} shows the results for the average time to correctly predict the robot's objective, with the standard error bars.

\begin{figure}[t]
	\centering
	\includegraphics[width=0.7\linewidth]{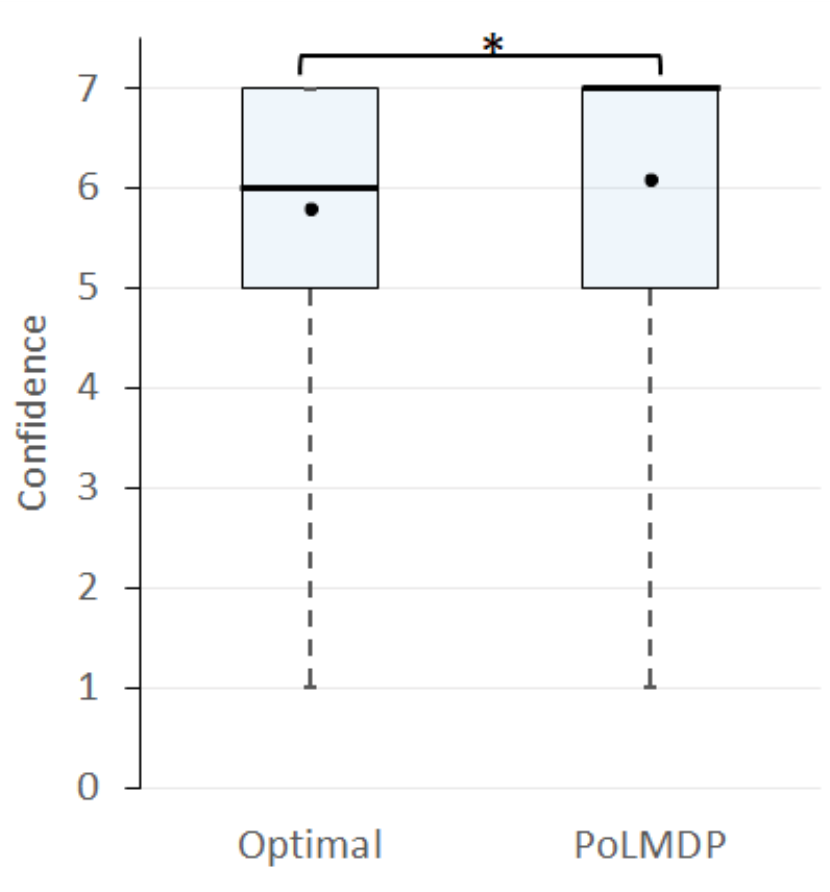}
	\caption{Boxplot comparing the self-rated confidences in the predictions, according to the type of policy used. The average for each policy is marked with a dot and thicker black line marks the median. (*$p < 0.05$)}
	\label{fig:study-confidence}
\end{figure}

Finally, regarding the self-rated confidence in the predictions, Figure~\ref{fig:study-confidence} shows a boxplot comparing the two conditions. Participants paired with the \ac{PoLMDP} condition rated their confidence, on average, as $6.08$ out of $7$ while participants paired with the optimal condition rated as $5.78$ out of $7$. A Mann-Whitney test conducted compared the two conditions averages, $U = 231203, p = 0.001$, showing that participants paired with the \ac{PoLMDP} condition were statistically more confident in their predictions than those paired with the optimal condition. These results support hypothesis \textbf{H2}.

\subsubsection{Discussion}
\label{subsubsec:study-discussion}

The results of the user study support both of our working hypotheses, thus showing the positive impact of using our \ac{PoLMDP} policy. Our approach allowed the robot to be more expressive regarding its internal goals, making clearer for the users interacting with the robot what the robot was trying to achieve.

Participants paired with our legible policy saw an increase of ~15\% in correct predictions, taking on average ~$3$ seconds less to predict the robot's objective. These two aspects support the usefulness of this type of policy in interaction scenarios between humans and robots: by causing humans to have better predictions about a robot's intentions and with less time needed to predict, humans have more time to analyse the workspace and decide on the best action to perform according to their own internal objectives and intentions. Thus, humans can think about what they are doing, instead of just reacting to a robot, giving humans more power in the interaction.

Another aspect to highlight from this study is the higher confidence human participants felt in their predictions. The confidence was not only higher, but consistently higher in humans paired with \ac{PoLMDP} as seen by the boxplot in Figure~\ref{fig:study-confidence} where the median line is at the highest possible value. This consistency of high confidence is important, because when humans feel confident in their decisions they can focus better in their personal objectives and tasks and, in collaborative scenarios, this can lead to increased task performance. Thus, an increase in confidence has a great impact in the success of an interaction and the overall interaction experience.

\section{Conclusion}
\label{sec:conclusion}


In this paper we present \ac{PoLMDP}, a framework for sequential decision making, based on \acp{MDP}, that creates legible behaviours, aimed at improving an intelligent system's expressiveness. \ac{PoLMDP} generates legible actions and behaviours, by choosing the actions that at a given moment in time are more representative of the robot's intentions -- distancing the robot from other possible objectives -- and giving humans more insights into the robot's internal state.

Through a combination of two simulation evaluations and one online user study, we have shown the positive impact of our \ac{PoLMDP} framework. In a simulation of performance with a similar framework, we showed that \ac{PoLMDP} can outperform it, generating legible behaviours in less time allowing agents to be more efficient. In a simulation with an optimal agent, we showed that the examples given by \ac{PoLMDP} are better at showing the task being executed, allowing for an \ac{IRL} agent to learn the underlying reward functions faster than with the optimal agent. Finally, through an online user study, we showed that \ac{PoLMDP} is better at generate behaviours that convey an autonomous system's internal state than using optimal policies, giving humans more time to adapt to a system's actions and act accordingly.

We envision \ac{PoLMDP} as a promising framework for applications in scenarios of search and rescue operations, where a team must coordinate to explore an area with various points of interest; for collaborative applications in healthcare scenarios where reading actions from your co-workers body language is essential for team coordination; or for collaboration tasks in home scenarios, where the robot acts in an independent capacity to achieve a set of objectives -- like a roomba cleaning the house.


\bibliographystyle{unsrtnat}
\bibliography{references}






\end{document}